\documentclass[10pt]{article}
\usepackage[preprint]{tmlr}

\usepackage{amsmath, amssymb, amsfonts}
\usepackage{booktabs}
\usepackage{multirow}
\usepackage{graphicx}
\usepackage{url}
\usepackage{hyperref}
\usepackage{xcolor}
\usepackage{enumitem}
\usepackage{array}
\usepackage{caption}
\usepackage{float}
\usepackage{tabularx}
\usepackage{makecell}
\usepackage{placeins}
\newcommand{\fallbackfigure}[2]{%
\IfFileExists{#1}{\includegraphics[width=\linewidth]{#1}}{%
\fbox{\parbox{0.92\linewidth}{\centering \vspace{0.6em}#2\\[0.6em]Figure file \texttt{\detokenize{#1}} not found in this draft package.\vspace{0.6em}}}}}

\title{Forecasting Is Not Attribution:\\Localizing Decoder Bypass in Graph-Based Neural Marketing Mix Models}

\author{\name Yunbo Wang \email yunbow4@uci.edu \\
      \addr Department of Electrical Engineering and Computer Science\\
      University of California, Irvine
      \AND
      \name Bolbi Liu \email bolbi@adsgency.ai \\
      \addr AdsGency AI
}

\def\month{MM}
\def\year{YYYY}
\date{}

\newcommand{\dice}{DICE-MMM}
\newcommand{\causalmmm}{CausalMMM}
\newcommand{\cig}{CIG}
\newcommand{\arcig}{AR-CIG}
\newcommand{\E}{\mathbb{E}}
\newcommand{\R}{\mathbb{R}}
\newcommand{\Z}{\mathbf{Z}}
\newcommand{\X}{\mathbf{X}}
\newcommand{\Y}{\mathbf{Y}}
\newcommand{\V}{\mathbf{V}}
\newcommand{\cctx}{\mathbf{c}}

\begin{document}

\maketitle

\begin{abstract}
Marketing mix models are used both to forecast business outcomes and to attribute those outcomes to marketing channels. These two goals are not equivalent. We study a failure mode in graph-based neural MMM, which we call \emph{attribution bypass}: a high-capacity decoder can obtain low forecasting error through target autoregression, dense communication, channel co-movement, context, or latent memory while failing to route its counterfactual sensitivity through the graph later used as the attribution object.

This paper makes a deliberately bounded claim. We do not claim that observational neural MMM identifies causal effects. Instead, we introduce \dice{} as a graph-learning and graph-use diagnostic framework that separates three questions often conflated in graph-based MMM: whether the model recovers a plausible temporal graph, whether it forecasts accurately, and whether the trained decoder's perturbation-induced influence is graph aligned. DICE Stage 1 trains a graph encoder with a restricted graph-mediated decoder. Stage 2 freezes the selected encoder and trains a graph-safe latent decoder whose cross-node communication must pass through the supplied graph. We evaluate decoder use with counterfactual influence graphs (\cig), autoregressive rollout influence graphs (\arcig), and frozen-decoder graph-swap tests.

The empirical picture is intentionally not overclaimed. DICE improves stable graph recovery over \causalmmm{} across controlled $R/d/T$ swaps and an external multi-graph rawlog stress test. However, forecasting accuracy is not an attribution certificate: in a non-degenerate sparse-target benchmark, no-graph and full-graph decoders achieve MSE@7 around $0.004$ while AR-CIG nAUPRC remains near or below zero, whereas an oracle graph achieves AR-CIG nAUPRC $0.807\pm0.129$ at comparable MSE. Frozen graph-swap further shows that the decoder and AR-CIG diagnostic are functional: the same DICE-hard-trained decoder moves from nAUPRC $-0.044\pm0.006$ under hard/raw/full learned graph inputs to $0.894\pm0.027$ when supplied with the oracle graph. The remaining failure is also clear. Learned DICE graph interfaces and label-free sparsification rules such as validation-MSE selection and stability selection remain insufficient. The contribution is therefore a stress test and failure-localization framework: it proves that low MSE can hide attribution bypass and localizes the unsolved bottleneck to deployable graph-support selection rather than to forecasting or decoder capacity.
\end{abstract}

\section{Introduction}

Marketing mix modeling (MMM) is a decision tool, not merely a forecasting tool\citep{Borden1964TheCO,Gujar2024TheEO}. A deployed MMM system is used to decide which channels deserve budget, which channels should be reduced, and which historical changes would have mattered under a different allocation. These decisions require attribution. Yet many neural MMM evaluations validate only prediction error. This creates a dangerous shortcut: a model can forecast well by exploiting target history, shared seasonality, channel co-movement, dense communication, or latent memory, without learning a graph that supports faithful attribution.

Graph-based neural MMM tries to expose the attribution interface by learning a directed temporal graph and forecasting through it. This is appealing only if the decoder actually uses the graph. A strong decoder can route information around a weak or dense graph and still reduce held-out MSE. We call this failure mode \emph{attribution bypass}. It is especially problematic in MMM because a low validation error can make an unfaithful attribution graph look validated.

This paper is written to withstand the strictest interpretation of that concern. We do not present \dice{} as a solved causal-attribution method. We present it as a diagnostic framework for separating graph recovery, forecasting, and decoder-induced graph use. DICE Stage 1 trains a graph encoder with a restricted graph-mediated decoder, so the graph-discovery gradients are not immediately dominated by a high-capacity response model. DICE Stage 2 freezes the selected encoder and trains a graph-safe latent decoder, so response modeling cannot rewrite the graph. To test whether the supplied graph matters, we measure decoder-induced influence under source perturbations and perform frozen-decoder graph swaps.

The main empirical result is a separation. In the sparse-target benchmark, full and no-graph decoders forecast about as well as oracle-graph decoders, but their AR-CIG alignment is near chance. Oracle graph support yields high AR-CIG. More importantly, the same trained decoder produces near-chance AR-CIG under dense learned graph inputs and high AR-CIG when swapped to the oracle graph. Thus the decoder is not simply graph blind, and the AR-CIG diagnostic is not vacuous. The bottleneck is the learned graph-to-support interface: current raw/hard posterior interfaces are too dense, and simple label-free selectors are not yet reliable.

This leads to a conservative but useful conclusion. \dice{} does not show that a learned neural MMM graph is already a deployable attribution object. It shows how to detect when it is not, how to rule out several misleading explanations, and where the remaining technical problem lies.

\paragraph{Contributions.}
\begin{enumerate}[leftmargin=2em]
    \item We identify \emph{attribution bypass}, a failure mode in graph-based neural MMM where low forecasting error coexists with decoder-induced influence that is not aligned with the attribution graph.
    \item We propose \dice{}, a two-stage graph-learning and diagnostic framework that protects the graph as an interface between discovery and high-capacity response modeling.
    \item We define CIG/AR-CIG diagnostics and frozen graph-swap tests to distinguish three failure explanations: a broken metric, a graph-blind decoder, and a weak learned graph support.
    \item We provide empirical failure localization. Oracle graphs produce high AR-CIG, and the same frozen decoder responds correctly when swapped to the oracle graph, but current learned graph interfaces and label-free sparsification remain insufficient. This localizes the open problem to deployable sparse graph-support selection.
\end{enumerate}

\begin{table}[H]
\centering
\small
\setlength{\tabcolsep}{4pt}
\renewcommand{\arraystretch}{1.12}
\caption{\textbf{Claim--evidence map.} The manuscript is framed around claims that the current experiments can support. The boundary column states what is \emph{not} being claimed.}
\label{tab:claim_evidence}
\begin{tabularx}{\textwidth}{p{0.25\textwidth}p{0.45\textwidth}p{0.23\textwidth}}
\toprule
Claim & Evidence & Boundary \\
\midrule
DICE improves graph recovery. & Table~\ref{tab:graph_recovery} shows higher stable Final-20 AUROC than \causalmmm{} across controlled $R/d/T$ swaps; Table~\ref{tab:rawlog} shows the same direction in an external rawlog stress test. & Graph recovery alone does not certify attribution. \\
Low MSE is not an attribution certificate. & Tables~\ref{tab:sparse_target} and~\ref{tab:bypass} show no/full/random or dense graph decoders with competitive MSE but near-zero or negative AR-CIG nAUPRC. & This is a controlled graph-known diagnostic, not a real-world causal proof. \\
The decoder and AR-CIG diagnostic are functional under correct graph support. & Oracle graph rows reach AR-CIG nAUPRC $0.807\pm0.129$ in sparse-target retraining and $0.894\pm0.027$ in frozen graph-swap. & Oracle support is an upper bound, not a deployable learned graph. \\
The decoder is not simply ignoring graph input. & In Table~\ref{tab:graphswap}, the same DICE-hard-trained decoder changes from $-0.044\pm0.006$ nAUPRC under hard/raw/full graph inputs to $0.894\pm0.027$ under oracle graph input. & This does not imply the learned graph support is sufficient. \\
Current learned graph interfaces remain insufficient. & Tables~\ref{tab:interface_selectors} and~\ref{tab:sparse_target} show validation-MSE sparsification, stability selection, and true-density top-$k$ only modestly improve over dense/no/full controls and remain far below oracle. & The paper localizes the bottleneck; it does not solve deployable support selection. \\
\bottomrule
\end{tabularx}
\end{table}

\paragraph{What this paper does not claim.}
We do not claim that CIG or AR-CIG is a causal estimand. We do not claim that observational graph learning identifies marketing causal effects without temporal, experimental, and confounding assumptions. We do not claim that the current learned DICE graph is already a deployable attribution graph. The strongest supported claim is narrower: forecasting accuracy, graph recovery, and decoder-induced attribution faithfulness can sharply diverge, and the proposed diagnostics can localize the divergence.

\section{Related Work}

\paragraph{Marketing mix modeling.}
Classical and Bayesian MMM models incorporate carryover, adstock, and nonlinear shape effects to estimate channel response and support budget allocation decisions \citep{46001}. Recent graph-based MMM extends this line by learning heterogeneous channel structures and response patterns from data. The closest baseline is \causalmmm{}, which formulates causal MMM as a graph variational autoencoder with a causal relational encoder and a marketing response decoder \citep{causalmmm}. Our work keeps the graph-based MMM motivation but asks a different question: after a graph is learned, does the final decoder actually use it for attribution?

\paragraph{Temporal causal discovery and Granger-style graphs.}
Granger causality interprets temporal influence as predictive usefulness of one variable's past for another variable's future under conditioning \citep{702ab909-8cb1-3c30-a5f1-ab4517d6cf1c}. Neural Granger methods extend this idea with sparse neural predictors \citep{Tank_2021}; PCMCI and DYNOTEARS provide alternative time-series causal discovery approaches based on conditional independence testing and score-based dynamic Bayesian network learning \citep{Runge_2019,pamfil2020dynotearsstructurelearningtimeseries}. These methods help define and recover temporal dependence structures, but a recovered graph is not automatically a faithful explanation of a separate high-capacity decoder. DICE focuses on this decoder-interface gap.

\paragraph{High-capacity decoders and explanation faithfulness.}
Transformers and latent-memory decoders can model long-range temporal patterns and nonlinear interactions \citep{vaswani2017attention,fang2026rethinkingzeroshottimeseries,bailie2025hierarchicalgraphnetworksaccurate}. Their flexibility is useful for forecasting, but it also creates shortcut routes around explicit attribution interfaces. Explanation literature has similarly warned that model-internal weights such as attention are not guaranteed to be faithful explanations of model behavior \citep{jain2019attentionexplanation}. We adapt this concern to neural MMM: the relevant explanation object is not only a learned graph, but whether the trained decoder's counterfactual sensitivity is mediated by that graph.

\section{Problem Setup}

\subsection{Multivariate MMM Time Series}

We observe $N$ entities, such as shops, brands, regions, or campaign groups. For entity $n$, let
\begin{equation}
    \V_{n,t} = [X_{n,t,1},\ldots,X_{n,t,d}, y_{n,t}] \in \R^{C}, \qquad C=d+1,
\end{equation}
where $X_{n,t,i}$ is the spend, exposure, or activity of marketing variable $i$, and $y_{n,t}$ is the target KPI. Each entity also has optional context $\cctx_n$. The forecasting task is to predict a future horizon
\begin{equation}
    \hat y_{n,t+1:t+M} = F_\theta(\V_{n,1:t},\cctx_n,\Z_n),
\end{equation}
where $\Z_n$ is an entity-specific directed temporal graph.

In our controlled synthetic benchmark, the graph label is known. We use the directed off-diagonal edge set
\begin{equation}
    \mathcal{E}=\{(i,j): i,j\in\{1,\ldots,C\}, i\neq j\}, \qquad |\mathcal{E}|=C(C-1),
\end{equation}
and store labels as edge vectors. We follow the source-to-receiver convention: $A_{ij}=1$ means variable $i$ influences variable $j$. Self-links are excluded from graph recovery metrics.

\subsection{Granger-Style Graph Interpretation}

We interpret $Z_{n,ij}=1$ as a Granger-style temporal influence: the past trajectory of source variable $i$ is predictively useful for receiver variable $j$ after conditioning on the observed multivariate history. On synthetic data, $Z_n$ is compared to the ground-truth data-generating graph. On observational data, $Z_n$ should be read as a structural predictive graph unless stronger assumptions hold: temporal precedence, adequate variation, observed confounder control, and stable response mechanisms.

\subsection{Attribution Bypass}

A high-capacity decoder can be accurate for the wrong reason. Let $\mathcal{L}_{\mathrm{pred}}(F)$ denote forecasting loss on the observational distribution:
\begin{equation}
    \mathcal{L}_{\mathrm{pred}}(F)=\E_{\V\sim P_{\mathrm{obs}}}\left[\|F(\V)-\Y\|_2^2\right].
\end{equation}
Prediction loss only constrains the model on observed trajectories. Attribution probes evaluate the model under perturbed trajectories, which may lie outside the observational support. Therefore two decoders can have identical observational forecasting loss but different counterfactual behavior.

\paragraph{Diagnostic proposition.}
Let $\mathcal{M}$ be the support of observed trajectories. Suppose two decoders $F_1$ and $F_2$ satisfy $F_1(\V)=F_2(\V)$ for all $\V\in\mathcal{M}$. For a perturbation operator $B_i$ that modifies source variable $i$, $B_i(\V)$ need not lie in $\mathcal{M}$. Thus $F_1(B_i(\V))$ and $F_2(B_i(\V))$ can differ even when $F_1$ and $F_2$ have the same forecasting error. Forecasting accuracy alone therefore cannot certify attribution faithfulness.

\subsection{Decoder-Induced Counterfactual Influence Graph}

We define the decoder-induced counterfactual influence graph (CIG) as a diagnostic for attribution bypass. CIG is not a causal estimand. It measures how the trained decoder's prediction changes when one source variable is perturbed.

Let $B_i^m(\V)$ be a perturbation of source variable $i$ under mode $m\in\{\mathrm{zero},\mathrm{mean},\mathrm{shuffle}\}$. Zero replacement removes the source history, mean replacement substitutes the training-set node mean, and shuffle replacement swaps the source trajectory across entities. For a trained decoder $F_\theta$, the one-step CIG score from source $i$ to receiver $j$ is
\begin{equation}
    S^{\mathrm{1step}}_{ij}=\E_{n,t}\left[\left|F_{\theta,j}(\V_{n,1:t})-F_{\theta,j}(B_i^m(\V_{n,1:t}))\right|\right].
\end{equation}
For forecasting, perturbations can propagate autoregressively. We therefore define
\begin{equation}
    S^{\mathrm{AR}}_{ij}=\E_{n,t,h}\left[\left|\hat V^{\mathrm{AR}}_{n,t+h,j}(\V)-\hat V^{\mathrm{AR}}_{n,t+h,j}(B_i^m(\V))\right|\right], \qquad h=1,\ldots,M.
\end{equation}
AR-CIG perturbs only the observed prefix and compares the rolled-out forecasts over the final horizon. We convert CIG scores to edge rankings and report AUROC, AUPRC, and normalized AUPRC against the true graph on synthetic data.

\section{DICE-MMM}

\subsection{Overview}

\dice{} is a two-stage graph-mediated MMM framework. It consists of a DICE graph encoder $q_\phi(\Z|\X)$ and a graph-safe latent response decoder $p_\psi(\Y|\X,\cctx,\Z)$. The design is intentionally asymmetric. Stage 1 is optimized for graph discovery: the encoder is trained with a restricted graph-mediated decoder so that predictive information must pass through candidate edges. Stage 2 is optimized for response modeling: the selected encoder is frozen and a stronger latent decoder is trained under the fixed graph. The central invariant is that the final decoder may be high-capacity in time and within-node latent memory, but every cross-node path must be mediated by the learned graph.

This separation is the meaning of \emph{decoder invariance}. The graph should not depend on whether the final response model is an MLP, an RNN, a transformer, or a latent-memory decoder. If the graph changes whenever the decoder becomes stronger, then the graph is not an attribution object; it is merely a byproduct of a forecasting architecture. \dice{} therefore treats the graph as a protected interface between discovery and forecasting.

\subsection{DICE Graph Encoder}

The encoder parameterizes a factorized categorical posterior over directed off-diagonal edges:
\begin{equation}
    q_\phi(\Z|\X)=\prod_{(i,j)\in\mathcal{E}} q_\phi(z_{ij}|\X), \qquad q_\phi(z_{ij}|\X)=\mathrm{Cat}(z_{ij};\pi_{ij}).
\end{equation}
It returns edge logits $a_{ij}\in\R^2$ for no-edge and edge classes, with $\pi_{ij}=\mathrm{softmax}(a_{ij})$. The interface remains compatible with \causalmmm-style graph-VAE training: the encoder returns $[B,E,2]$ edge logits, which can be sampled with Gumbel-softmax during Stage 1 and converted into soft or hard graph inputs during Stage 2.

The DICE encoder adds two graph-discovery biases. First, it uses strictly causal temporal node encodings: each node history is projected and passed through left-padded dilated residual temporal convolutions, followed by attention pooling and summary statistics. This prevents the encoder from using future observations when constructing node representations. Second, it constructs directed pair features from sender and receiver summaries, including sender--receiver contrasts, elementwise interactions, and lag-aware past(source) $\rightarrow$ future(receiver) correlations. For lags $\mathcal{D}$, a directed lag feature is
\begin{equation}
    \rho_d(i\to j)=\frac{1}{T-d}\sum_{t=1}^{T-d} \bar x_{t,i}\bar x_{t+d,j}, \qquad d\in\mathcal{D},
\end{equation}
combined with antisymmetric contrasts $\rho_d(i\to j)-\rho_d(j\to i)$. These features provide a lightweight Granger-style inductive bias without using ground-truth edge labels.

\begin{figure}[t]
    \centering
    \fallbackfigure{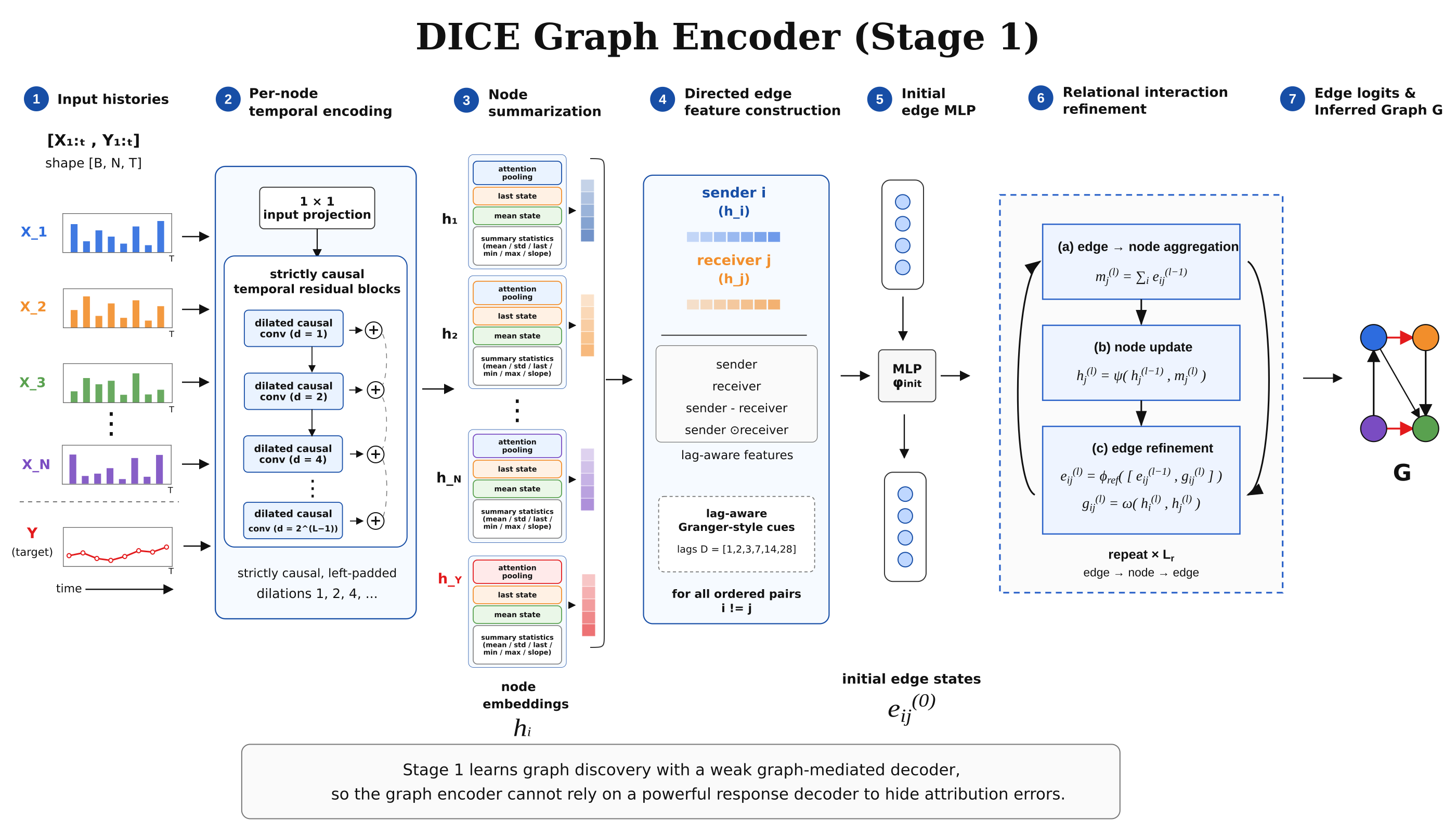}{DICE graph encoder schematic placeholder.}
    \caption{\textbf{DICE graph encoder used in Stage 1.} The encoder maps node histories to strictly causal temporal representations, summarizes each node, constructs directed sender--receiver features with lag-aware Granger-style cues, initializes edge logits with an MLP, and refines the edge states through edge--node--edge relational interaction. Stage 1 pairs this encoder with a restricted graph-mediated decoder so that graph discovery cannot be hidden by a high-capacity response decoder.}
    \label{fig:dice_encoder}
\end{figure}

Figure~\ref{fig:dice_encoder} shows the implemented encoder pipeline. The important architectural point is that DICE is not simply a larger encoder. Its temporal blocks make the node summaries causal; its lag features make the edge candidates directional; and its relational refinement lets a candidate edge be judged in the context of other incoming evidence. The output remains ordinary edge logits, so the same graph metrics, Gumbel sampling, and KL terms used by \causalmmm{} still apply.

\subsection{Stage 1: Restricted-Decoder Graph Discovery}

Stage 1 trains the graph encoder with a restricted graph-mediated decoder $f_{\theta_0}$. We sample a relaxed graph using Gumbel-softmax:
\begin{equation}
    \tilde z_{ij,k}=\frac{\exp((a_{ij,k}+g_{ij,k})/\tau)}{\sum_{k'\in\{0,1\}}\exp((a_{ij,k'}+g_{ij,k'})/\tau)}, \qquad g_{ij,k}\sim\mathrm{Gumbel}(0,1).
\end{equation}
The Stage-1 objective is a graph-VAE loss:
\begin{equation}
    \mathcal{J}_{\mathrm{stage1}}(\phi,\theta_0)=\E_{q_\phi(\Z|\X)}[-\log p_{\theta_0}(\X_{2:T}|\X_{1:T-1},\cctx,\Z)] + \lambda\,\mathrm{KL}(q_\phi(\Z|\X)\|p(\Z)).
\end{equation}
The restricted decoder is not a performance concession; it is an identification device for the graph-learning problem. If a powerful response decoder is used during graph discovery, prediction gradients can improve MSE by exploiting shortcuts that do not require correct edges. In the controlled synthetic setting, ground-truth edges are used for evaluation and diagnostic checkpointing only; no edge cross-entropy, ranking, or Brier supervision is used in the main Stage-1 training loss. A deployable observational system would require label-free checkpointing and sparsification, which we report as a limitation rather than hide.

\subsection{Stage 2: Frozen Graph-Safe Latent Response Decoder}

After Stage 1, we select a graph encoder checkpoint by the synthetic validation graph diagnostic or by a pre-specified late-window rule, freeze the encoder, and train only the Stage-2 decoder. The Stage-2 decoder receives either the posterior mean graph $\bar\Z=\E_{q_{\phi^\star}(\Z|\X)}[\Z]$ or a deterministic hard graph. Its objective emphasizes target forecasting:
\begin{equation}
    \mathcal{J}_{\mathrm{stage2}}(\psi;\phi^\star)=\beta_{\mathrm{all}}\mathcal{L}_{\mathrm{all}} + \beta_y\mathcal{L}_y + \beta_M\mathcal{L}_{y,M},
\end{equation}
where $\mathcal{L}_{\mathrm{all}}$ is all-node MSE, $\mathcal{L}_y$ is target MSE over the full prediction sequence, and $\mathcal{L}_{y,M}$ is target MSE over the final forecasting horizon. The Stage-2 checkpoint is selected by validation target-horizon MSE. This stage answers the forecasting question without allowing forecasting gradients to rewrite the graph.

\subsection{Graph-Safe Latent Decoder}

The Stage-2 decoder features a high-capacity yet graph-safe architecture, drawing inspiration from Latent Thought Models (LTM) \citep{kong2025latentthoughtmodelsvariational} and the factored attention framework of NNN \citep{mulc2025nnn}. It converts frozen edge probabilities into an adjacency matrix $A$, restricting cross-node communication to graph message passing while independently applying temporal self-attention.
\begin{equation}
    \tilde h_{t,j}=h_{t,j}+\eta\frac{\sum_i A_{ij} W_g h_{t,i}}{\sum_i A_{ij}+\epsilon}.
\end{equation}
Temporal self-attention is applied independently per node, not over all time--node tokens. Node-specific causal latent memory slots are generated from graph-filtered prefixes:
\begin{equation}
    p_{t,j}=\frac{1}{t}\sum_{s=1}^t h_{s,j}, \qquad \zeta_{t,j,1:K}=m_\psi(p_{t,j},\cctx,d_j),
\end{equation}
where $d_j$ contains graph degree features. Each node attends only to its own latent slots. Thus, the target node cannot access a global latent memory that pools all channels and bypasses the graph. Optional context-conditioned Hill saturation can be applied to the raw target prediction to model diminishing returns.

\begin{figure}[t]
    \centering
    \fallbackfigure{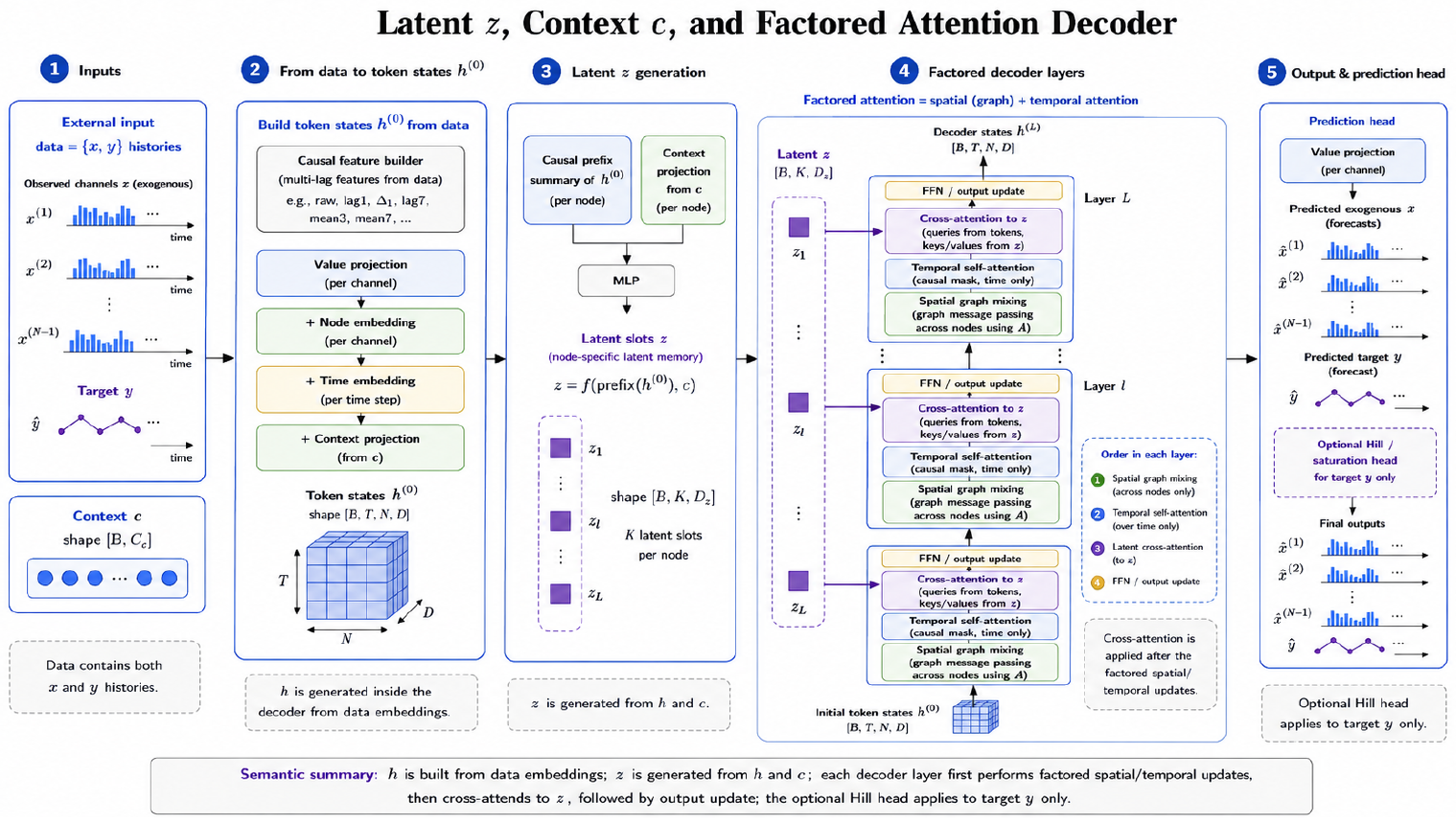}{Graph-safe latent decoder schematic placeholder.}
    \caption{\textbf{Graph-safe latent decoder used in Stage 2.} The decoder builds causal token states from historical data and context, generates node-specific latent memory slots $\zeta$, and applies factored graph--temporal--latent updates. Spatial communication is constrained by the frozen graph, temporal attention is per node, and latent cross-attention uses node-specific slots rather than a global memory that could bypass $\Z$.}
    \label{fig:latent_decoder_arch}
\end{figure}

Figure~\ref{fig:latent_decoder_arch} illustrates why Stage 2 can be expressive without destroying attribution. The latent decoder can model nonlinear carryover, delayed response, contextual heterogeneity, and temporal dependencies, but the only way one channel can affect another node's state is through the frozen graph adjacency. This is the mechanism that turns graph recovery into graph-mediated forecasting rather than merely reporting a graph next to a separate black-box forecaster.

\section{Experiments}

We organize the empirical section around six questions a strict reviewer should ask. Table~\ref{tab:graph_recovery} asks whether the DICE encoder improves graph recovery. Table~\ref{tab:forecasting} asks whether restricted graph discovery sacrifices forecasting. Table~\ref{tab:sparse_target} asks whether low MSE certifies attribution in a non-degenerate sparse-target benchmark. Table~\ref{tab:bypass} repeats the bypass diagnostic in the original response setting. Table~\ref{tab:graphswap} asks whether the decoder is graph-blind by swapping graph inputs while freezing decoder weights. Table~\ref{tab:interface_selectors} asks whether current learned graph interfaces and label-free sparsification solve the support-selection problem. Table~\ref{tab:rawlog} reports an external graph-recovery stress test.

\subsection{Experimental Protocol}

\paragraph{Data.}
We use controlled synthetic MMM datasets with known directed graphs, following the data generation protocol established by CausalMMM\citep{causalmmm}. Each sample is an entity-level time series with $d$ marketing nodes and one target node. The generator samples one of $R$ latent graph prototypes for each entity, then generates channel, mediator, and target trajectories with lagged nonlinear temporal dependencies and optional Hill-style saturation. Table~\ref{tab:graph_recovery} follows one-factor-at-a-time $R/d/T$ graph-recovery swaps. The original response-diagnostic setting uses $R=10,d=10,T=120$ and contains a degenerate target-parent label set, so attribution is evaluated over all directed off-diagonal edges. The new sparse-target benchmark removes that degeneracy by making only a subset of channels direct or mediated drivers of the target, allowing target-facing attribution diagnostics in addition to all-edge AR-CIG.

\begin{figure}[t]
    \centering
    \fallbackfigure{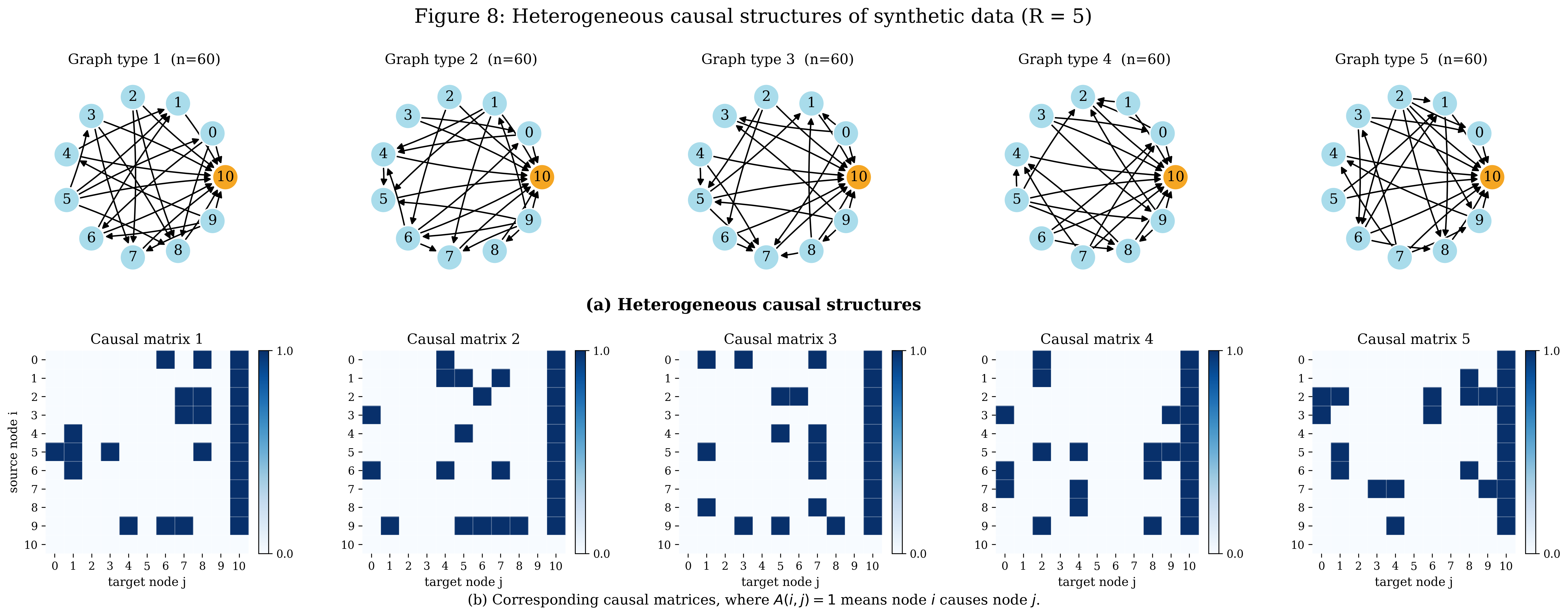}{Synthetic graph structures placeholder.}
    \caption{\textbf{Synthetic heterogeneous causal structures for the $R=5$ benchmark.} The top row shows five graph prototypes over ten channel nodes and one KPI node, and the bottom row shows the corresponding directed causal matrices. We use the convention $A(i,j)=1$ if source node $i$ influences receiver node $j$.}
    \label{fig:r5_synthetic_structures}
\end{figure}

\paragraph{Graph convention and metrics.}
We use the source-to-receiver convention: $A_{ij}=1$ means source node $i$ influences receiver node $j$. Self-links are excluded. Graph recovery is evaluated by AUROC. For Table~\ref{tab:graph_recovery}, we report \emph{Final-20} AUROC, the late-window average over the last 20 epochs, and \emph{Max} AUROC, the best validation AUROC diagnostic. Final-20 is the main checkpoint-stable number; Max is a sensitivity diagnostic.

Forecasting is evaluated by target MSE at fixed horizons. Attribution faithfulness is evaluated by AR-CIG AUPRC and normalized AUPRC,
\begin{equation}
    \mathrm{nAUPRC}=\frac{\mathrm{AUPRC}-\pi}{1-\pi},
\end{equation}
where $\pi$ is the edge prevalence. Thus $\mathrm{nAUPRC}\approx 0$ indicates prevalence-level attribution even when raw AUPRC is numerically nonzero. Negative nAUPRC indicates worse-than-prevalence ranking.

\paragraph{Seeds and checkpointing.}
All main tables report mean $\pm$ standard deviation over 10 pre-specified seeds unless otherwise stated. Stage-1 graph checkpoints are selected by the same late-window rule or validation graph diagnostic across graph-discovery methods in synthetic graph-known experiments. Stage-2 decoder checkpoints are selected by validation target-horizon MSE. Graph-swap experiments freeze decoder weights and change only the graph input at evaluation time. No seed is selected based on test performance.

\subsection{Graph Recovery under $R/d/T$ Swaps}

Table~\ref{tab:graph_recovery} evaluates whether DICE improves stable graph discovery. DICE Stage 1 improves Final-20 AUROC over CausalMMM in every completed column. This table supports the graph-recovery claim only; it does not imply that a downstream decoder uses the graph for attribution.

\begin{table}[H]
\centering
\small
\setlength{\tabcolsep}{4.2pt}
\renewcommand{\arraystretch}{1.10}
\caption{\textbf{Graph recovery under $R/d/T$ swaps.} We report AUROC mean $\pm$ std over 10 seeds. ``Final-20'' is the late-window average over the last 20 epochs and is the main checkpoint-stable number. ``Max'' is the best validation AUROC diagnostic and is treated as a sensitivity number. Bold indicates the higher value within each completed column.}
\label{tab:graph_recovery}

\textbf{(a) Channel-number swap: $R=5,T=120$}

\vspace{0.25em}
\begin{tabular}{lcccccc}
\toprule
& \multicolumn{2}{c}{$d=5$} & \multicolumn{2}{c}{$d=10$} & \multicolumn{2}{c}{$d=20$} \\
\cmidrule(lr){2-3}\cmidrule(lr){4-5}\cmidrule(lr){6-7}
Method & Final-20 & Max & Final-20 & Max & Final-20 & Max \\
\midrule
CausalMMM
& $0.672 \pm 0.239$ & $0.814 \pm 0.061$
& $0.688 \pm 0.058$ & $0.761 \pm 0.031$
& $0.611 \pm 0.007$ & $0.625 \pm 0.009$ \\
DICE-MMM Stage 1
& $\mathbf{0.726 \pm 0.019}$ & $\mathbf{0.823 \pm 0.032}$
& $\mathbf{0.707 \pm 0.020}$ & $\mathbf{0.762 \pm 0.018}$
& $\mathbf{0.614 \pm 0.030}$ & $\mathbf{0.653 \pm 0.029}$ \\
\bottomrule
\end{tabular}

\vspace{0.8em}
\textbf{(b) Series-length swap: $R=5,d=10$}

\vspace{0.25em}
\begin{tabular}{lcccccc}
\toprule
& \multicolumn{2}{c}{$T=30$} & \multicolumn{2}{c}{$T=120$} & \multicolumn{2}{c}{$T=720$} \\
\cmidrule(lr){2-3}\cmidrule(lr){4-5}\cmidrule(lr){6-7}
Method & Final-20 & Max & Final-20 & Max & Final-20 & Max \\
\midrule
CausalMMM
& $0.763 \pm 0.013$ & $0.777 \pm 0.015$
& $0.688 \pm 0.058$ & $0.761 \pm 0.031$
& $0.687 \pm 0.065$ & $0.779 \pm 0.027$ \\
DICE-MMM Stage 1
& $\mathbf{0.776 \pm 0.031}$ & $\mathbf{0.796 \pm 0.014}$
& $\mathbf{0.707 \pm 0.020}$ & $\mathbf{0.762 \pm 0.018}$
& $\mathbf{0.738 \pm 0.010}$ & $\mathbf{0.781 \pm 0.013}$ \\
\bottomrule
\end{tabular}

\vspace{0.8em}
\textbf{(c) Latent-structure-number swap: $d=10,T=120$}

\vspace{0.25em}
\begin{tabular}{lcccccc}
\toprule
& \multicolumn{2}{c}{$R=5$} & \multicolumn{2}{c}{$R=10$} & \multicolumn{2}{c}{$R=20$} \\
\cmidrule(lr){2-3}\cmidrule(lr){4-5}\cmidrule(lr){6-7}
Method & Final-20 & Max & Final-20 & Max & Final-20 & Max \\
\midrule
CausalMMM
& $0.688 \pm 0.058$ & $0.761 \pm 0.031$
& $0.643 \pm 0.065$ & $0.681 \pm 0.041$
& $0.580 \pm 0.073$ & $0.615 \pm 0.021$ \\
DICE-MMM Stage 1
& $\mathbf{0.707 \pm 0.020}$ & $\mathbf{0.762 \pm 0.018}$
& $\mathbf{0.668 \pm 0.024}$ & $\mathbf{0.752 \pm 0.009}$
& $\mathbf{0.612 \pm 0.080}$ & $\mathbf{0.702 \pm 0.055}$ \\
\bottomrule
\end{tabular}
\end{table}

\paragraph{Interpretation.}
DICE Stage 1 is consistently better under the stable Final-20 rule, but several improvements are modest. We therefore use this table only to establish that the DICE encoder carries graph-recovery signal. The later attribution tables decide whether that signal becomes decoder-induced attribution.

\subsection{Forecasting Sanity Check}

Table~\ref{tab:forecasting} tests whether restricted graph discovery comes at a prediction cost. It does not. DICE Stage 1 with a restricted graph-mediated decoder is already the strongest forecaster in this table. Therefore Stage 2 should not be justified as an MSE improvement; its purpose is to test graph-safe high-capacity response modeling under a frozen graph interface.

\begin{table}[H]
\centering
\small
\setlength{\tabcolsep}{5.2pt}
\renewcommand{\arraystretch}{1.12}
\caption{\textbf{Forecasting sanity check on the $R=10,d=10,T=120$ response-diagnostic setting.} All rows are selected by validation forecasting MSE under the same seed/split protocol. This table tests whether DICE graph learning incurs a prediction penalty. It is not evidence of attribution faithfulness.}
\label{tab:forecasting}
\resizebox{\textwidth}{!}{%
\begin{tabular}{llllccc}
\toprule
Method
& Graph source
& Final decoder
& Encoder in final stage
& MSE@1 $\downarrow$
& MSE@7 $\downarrow$
& MSE@30 $\downarrow$ \\
\midrule
CausalMMM original
& CREncoder
& MRDecoderS
& trained end-to-end
& $0.051 \pm 0.068$
& $0.253 \pm 0.212$
& $0.421 \pm 0.098$ \\

DICE-MMM Stage-1 only
& DICE encoder
& restricted graph-mediated MRDecoderS
& trained
& $\mathbf{0.011 \pm 0.006}$
& $\mathbf{0.129 \pm 0.097}$
& $\mathbf{0.238 \pm 0.062}$ \\

DICE-MMM Stage-2 freeze
& frozen DICE encoder
& LatentGraphDecoder
& frozen
& $0.058 \pm 0.104$
& $0.173 \pm 0.198$
& $0.244 \pm 0.282$ \\
\bottomrule
\end{tabular}}
\end{table}

\paragraph{Interpretation.}
The restricted decoder is restricted in bypass capacity, not in forecasting usefulness. This prevents a misleading explanation in which graph recovery improves only because prediction is sacrificed. It also reinforces the central point: forecasting sanity is necessary, but not sufficient for attribution.

\subsection{Non-Degenerate Sparse-Target Bypass Diagnostic}

Table~\ref{tab:sparse_target} is the cleanest attribution result. The sparse-target benchmark removes the degenerate target-parent issue from the original response-diagnostic setting. The decisive pattern is that no-graph and full-graph decoders achieve essentially the same MSE@7 as oracle-graph decoders, but their AR-CIG nAUPRC is near or below zero. Oracle graph support reaches high AR-CIG. This validates the diagnostic separation between prediction and graph-mediated influence.

\begin{table}[H]
\centering
\small
\setlength{\tabcolsep}{5.2pt}
\renewcommand{\arraystretch}{1.15}
\caption{\textbf{Sparse-target decoder diagnostic.} The sparse-target setting is non-degenerate for target-facing attribution. All rows use the same decoder family and report MSE@7 and AR-CIG nAUPRC. Low MSE alone is uninformative: no-graph and full-graph inputs forecast as well as oracle graph input but remain near prevalence in AR-CIG.}
\label{tab:sparse_target}
\resizebox{\textwidth}{!}{%
\begin{tabular}{lccc p{6.6cm}}
\toprule
Graph input / selector & Uses graph labels? & MSE@7 $\downarrow$ & AR-CIG nAUPRC $\uparrow$ & Interpretation \\
\midrule
Oracle graph & yes & $\mathbf{0.004 \pm 0.001}$ & $\mathbf{0.807 \pm 0.129}$ & Correct support yields graph-mediated influence at the same forecasting scale as shortcut controls. \\
Stability selection, freq $\geq 60\%$ & no & $0.004 \pm 0.002$ & $0.091 \pm 0.004$ & Modest positive signal, far below oracle. \\
Top-$k$ true density & yes & $0.005 \pm 0.002$ & $0.077 \pm 0.027$ & Diagnostic sparse support using oracle density; still far below oracle, so ranking is weak. \\
No graph & no & $\mathbf{0.004 \pm 0.001}$ & $-0.005 \pm 0.005$ & Temporal shortcut control: excellent MSE without graph-mediated influence. \\
Full graph & no & $\mathbf{0.004 \pm 0.001}$ & $-0.047 \pm 0.014$ & Dense communication control: excellent MSE but worse-than-prevalence attribution ranking. \\
\bottomrule
\end{tabular}}
\end{table}

\paragraph{Interpretation.}
This table is deliberately unfavorable to overclaiming. It proves that the decoder and AR-CIG can work under correct graph support, while showing that current learned supports do not approach that upper bound. It also gives the strongest version of the paper's central warning: in this benchmark, MSE@7 is essentially indistinguishable across oracle, no-graph, and full-graph inputs, while AR-CIG nAUPRC ranges from $0.807$ to negative values.

\subsection{Strong-Decoder Bypass Diagnostic in the Original Response Setting}

Table~\ref{tab:bypass} evaluates the same LatentGraphDecoder family under different graph inputs in the original $R=10,d=10,T=120$ response-diagnostic setting. Because target-parent labels are degenerate in this setting, we report all-edge AR-CIG. The oracle graph again yields high AR-CIG, while no/full/random graph controls forecast competitively with near-prevalence AR-CIG.

\begin{table}[H]
\centering
\small
\setlength{\tabcolsep}{3.7pt}
\renewcommand{\arraystretch}{1.12}
\caption{\textbf{Strong-decoder bypass diagnostic in the original response setting.} All rows use the same LatentGraphDecoder family; only the graph input changes. Low MSE is not sufficient for faithful attribution: full/no/random graph inputs can forecast competitively while producing chance-level AR-CIG.}
\label{tab:bypass}
\resizebox{\textwidth}{!}{%
\begin{tabular}{lccccp{5.2cm}}
\toprule
Model / graph input
& Graph AUROC $\uparrow$
& MSE@7 $\downarrow$
& AR-CIG AUPRC $\uparrow$
& AR-CIG nAUPRC $\uparrow$
& Interpretation \\
\midrule
DICE-MMM learned graph + latent decoder
& $\mathbf{0.750 \pm 0.009}$
& $0.334 \pm 0.247$
& $0.199 \pm 0.006$
& $-0.036 \pm 0.008$
& Better learned graph and MSE than the CausalMMM-graph hybrid, but all-edge AR-CIG remains weak. \\

CausalMMM graph + latent decoder
& $0.627 \pm 0.112$
& $0.521 \pm 0.264$
& $0.239 \pm 0.038$
& $0.016 \pm 0.049$
& Weaker learned graph; CIG remains near prevalence. \\

Oracle graph + latent decoder
& oracle
& $0.428 \pm 0.238$
& $\mathbf{0.903 \pm 0.024}$
& $\mathbf{0.874 \pm 0.031}$
& Attribution upper bound; validates the AR-CIG diagnostic. \\

No graph + latent decoder
& --
& $0.337 \pm 0.261$
& $0.222 \pm 0.010$
& $-0.007 \pm 0.013$
& Temporal shortcut control: competitive MSE without graph-mediated attribution. \\

Full graph + latent decoder
& --
& $\mathbf{0.305 \pm 0.223}$
& $0.192 \pm 0.005$
& $-0.046 \pm 0.007$
& Dense communication control: lowest MSE but poor attribution alignment. \\

Random graph + latent decoder
& random
& $0.434 \pm 0.277$
& $0.213 \pm 0.009$
& $-0.018 \pm 0.011$
& Sparse wrong-graph control; CIG collapses to chance. \\
\bottomrule
\end{tabular}}
\end{table}

\paragraph{Interpretation.}
The full graph has the best MSE@7 but negative AR-CIG nAUPRC. The oracle graph does not have the best MSE@7 but has high AR-CIG alignment. Thus observational forecasting error cannot validate attribution. The learned DICE graph improves graph AUROC, but the hard/raw learned-graph interface remains near chance. This motivates the graph-swap and graph-interface analyses below.

\subsection{Frozen-Decoder Graph Swap}

Table~\ref{tab:graphswap} is the strongest failure-localization experiment. It freezes decoder weights and changes only the graph input. If the decoder were completely graph blind, swapping to the oracle graph would not change AR-CIG. Instead, the same DICE-hard-trained decoder jumps from negative AR-CIG nAUPRC under hard/raw/full graph inputs to $0.894\pm0.027$ under oracle graph input.

\begin{table}[H]
\centering
\small
\setlength{\tabcolsep}{5.3pt}
\renewcommand{\arraystretch}{1.14}
\caption{\textbf{Frozen-decoder graph swap.} Decoder weights are fixed; only the graph input is swapped. $\Delta$Pred MAE@7 measures the prediction change relative to the trained graph input and is diagnostic rather than an accuracy metric. The large AR-CIG jump under oracle graph input shows that the decoder can use a correct graph.}
\label{tab:graphswap}
\begin{tabular}{llccc}
\toprule
Fixed decoder & Evaluation graph & MSE@7 $\downarrow$ & $\Delta$Pred MAE@7 & AR-CIG nAUPRC $\uparrow$ \\
\midrule
DICE-hard trained & hard / raw / full & $0.336 \pm 0.244$ & $0.000 \pm 0.000$ & $-0.044 \pm 0.006$ \\
DICE-hard trained & oracle & $0.360 \pm 0.244$ & $0.036 \pm 0.018$ & $\mathbf{0.894 \pm 0.027}$ \\
DICE-hard trained & top-$k$ true density & $0.513 \pm 0.265$ & $0.080 \pm 0.032$ & $0.137 \pm 0.066$ \\
\midrule
Oracle-trained & oracle & $0.428 \pm 0.238$ & $0.000 \pm 0.000$ & $0.874 \pm 0.031$ \\
Oracle-trained & top-$k$ true density & $0.590 \pm 0.251$ & $0.063 \pm 0.026$ & $0.142 \pm 0.063$ \\
Oracle-trained & no / full / random & low / competitive & nonzero & near $0$ \\
\bottomrule
\end{tabular}
\end{table}

\paragraph{Interpretation.}
The graph-swap experiment rules out two misleading explanations. First, AR-CIG is not merely broken: oracle graph input yields high AR-CIG. Second, the latent decoder is not incapable of using graph input: the same frozen decoder becomes graph aligned when supplied with the oracle graph. The remaining failure is more specific: the learned graph support delivered to the decoder is too dense or too weakly ranked. The top-$k$ rows show modest transfer signal, but they remain far from oracle.

\subsection{Learned Graph Interfaces and Label-Free Selectors}

Table~\ref{tab:interface_selectors} asks whether the learned DICE posterior can be converted into deployable sparse graph support. The answer is currently no. Raw and hard interfaces collapse into dense support; true-density top-$k$ is diagnostic because it uses graph-label information; and two label-free selectors, validation-MSE sufficiency and stability selection, do not recover attribution-aligned support.

\begin{table}[H]
\centering
\small
\setlength{\tabcolsep}{4.2pt}
\renewcommand{\arraystretch}{1.13}
\caption{\textbf{Learned graph interfaces and graph-support selectors.} These experiments test whether learned edge scores can be converted into decoder-consumable sparse support. Label-free selectors are deployable in principle but remain insufficient in the current implementation. True-density top-$k$ is an oracle-density diagnostic, not a deployable method.}
\label{tab:interface_selectors}
\resizebox{\textwidth}{!}{%
\begin{tabular}{lllccc p{5.1cm}}
\toprule
Setting & Graph interface / selector & Uses graph labels? & Density / support & MSE@7 $\downarrow$ & AR-CIG nAUPRC $\uparrow$ & Interpretation \\
\midrule
Old R10/M7 & raw posterior & no & mean weight $1.000\pm0.000$ & $0.305\pm0.220$ & $-0.039\pm0.010$ & Posterior ranking exists, but soft interface is effectively dense. \\
Old R10/M7 & hard threshold $p(e)>0.5$ & no & $110.0\pm0.0$ edges & $0.336\pm0.244$ & $-0.044\pm0.006$ & Threshold selects all edges; equivalent to full graph. \\
Old R10/M7 & temperature soft, $\tau=5$ & no & mean weight $0.893\pm0.016$ & $0.322\pm0.195$ & $-0.043\pm0.006$ & Softer probabilities remain too dense. \\
Old R10/M7 & validation-MSE sparse, $\epsilon=3\%$ & no & selected & $0.297\pm0.232$ & $-0.028\pm0.021$ & MSE-sufficient sparsification fails as attribution selection. \\
Old R10/M7 & stability selection, freq $\geq60\%$ & no & selected & $0.265\pm0.217$ & $-0.048\pm0.011$ & Stable edges are not necessarily attribution-faithful edges. \\
Old R10/M7 & top-$k$ true density & yes & $25.0\pm0.0$ edges & $0.417\pm0.335$ & $0.132\pm0.064$ & Oracle-density sparsification gives modest lift, not a deployable solution. \\
Old R10/M7 & top-$k$ high precision & partly diagnostic & $13.0\pm0.0$ edges & $0.315\pm0.236$ & $0.080\pm0.048$ & Higher precision support gives a small positive signal. \\
Old R10/M7 & one-stage strong end-to-end & no & learned & $\mathbf{0.158\pm0.168}$ & $-0.051\pm0.011$ & Lowest MSE, but graph AUROC collapses and AR-CIG is negative. \\
Sparse-target & stability selection, freq $\geq60\%$ & no & selected & $0.004\pm0.002$ & $0.091\pm0.004$ & Positive but far below oracle $0.807\pm0.129$. \\
Sparse-target & top-$k$ true density & yes & true density & $0.005\pm0.002$ & $0.077\pm0.027$ & Even oracle-density top-$k$ remains weak, showing edge ranking is not sufficient. \\
\bottomrule
\end{tabular}}
\end{table}

\paragraph{Interpretation.}
This table is the main reason the manuscript must be framed as diagnostic rather than as a solved attribution method. Validation-MSE selection fails because MSE is itself the wrong certificate. Stability selection fails because stable predictive edges can reflect seasonality, autoregression, or channel co-movement rather than target-relevant attribution structure. True-density top-$k$ and high-precision top-$k$ show only modest transfer signal. Therefore the remaining problem is not merely choosing a threshold; it is learning or selecting sparse graph support that is faithful to decoder-induced target influence.

\subsection{External Multi-Graph Rawlog Stress Test}

Table~\ref{tab:rawlog} tests whether the DICE graph encoder advantage is specific to the controlled synthetic generator. This is an external graph-recovery stress test, not an attribution table. DICE rawlog with magnitude prior substantially improves Final-20 AUROC over CausalMMM original rawlog. The Hill/saturation toggle has little effect on DICE graph recovery, suggesting that the gain comes primarily from the graph encoder and magnitude prior rather than from the saturation head.

\begin{table}[H]
\centering
\small
\setlength{\tabcolsep}{6pt}
\renewcommand{\arraystretch}{1.12}
\caption{\textbf{External multi-graph rawlog stress test.} We report graph AUROC on the CausalRiver/rawlog multi-graph setting. This table evaluates graph recovery robustness only; it is not a decoder-induced attribution experiment.}
\label{tab:rawlog}
\begin{tabular}{llcc}
\toprule
Model
& Hill/saturation
& Final-20 AUROC $\uparrow$
& Max AUROC $\uparrow$ \\
\midrule
DICE rawlog + magnitude prior
& off
& $0.6607 \pm 0.0163$
& $\mathbf{0.6692 \pm 0.0019}$ \\
DICE rawlog + magnitude prior
& on
& $\mathbf{0.6632 \pm 0.0097}$
& $\mathbf{0.6692 \pm 0.0019}$ \\
CausalMMM original rawlog
& on
& $0.4557 \pm 0.0112$
& $0.6115 \pm 0.0223$ \\
CausalMMM original rawlog
& off
& $0.4281 \pm 0.0081$
& $0.6128 \pm 0.0212$ \\
\bottomrule
\end{tabular}
\end{table}

\paragraph{Interpretation.}
The external stress test supports the graph-discovery claim from Table~\ref{tab:graph_recovery}. It should not be used to support attribution faithfulness. The paper's attribution evidence comes from oracle/no/full/random controls, sparse-target diagnostics, and frozen graph-swap.

\section{Discussion}

The experiments support a narrower and stronger story than ``DICE-MMM is a better forecaster.'' They separate four claims.

First, DICE improves graph discovery. Table~\ref{tab:graph_recovery} shows stronger stable graph recovery under $R/d/T$ swaps, and Table~\ref{tab:rawlog} shows the same direction in an external multi-graph rawlog stress test. This establishes that the encoder is not random, but graph recovery is not attribution.

Second, forecasting and attribution are empirically separable. Tables~\ref{tab:sparse_target} and~\ref{tab:bypass} show that no-graph and full-graph decoders can forecast competitively while AR-CIG remains near prevalence. The sparse-target result is particularly stark: oracle, no-graph, and full-graph inputs all obtain MSE@7 around $0.004$, but only the oracle graph yields high AR-CIG alignment.

Third, the AR-CIG diagnostic and graph-safe decoder are functional under correct graph input. Oracle graph support yields high AR-CIG when the decoder is trained with it, and the frozen graph-swap experiment shows that the same decoder can become graph aligned when supplied with the correct graph. This rules out a simple ``the decoder ignores graphs'' explanation.

Fourth, the current learned graph interface is insufficient. Raw posterior and hard-threshold interfaces are too dense. Validation-MSE sparsification fails because MSE is not an attribution certificate. Stability selection gives only a small lift in the sparse-target setting and fails in the original response setting. Top-$k$ true density gives modest positive AR-CIG but is diagnostic because it uses oracle density. The remaining technical challenge is therefore deployable graph-support selection: converting learned temporal edge scores into sparse support that is both predictive enough and attribution aligned.

This is the central contribution of the paper. The experiments do not prove that DICE has solved neural MMM attribution. They prove that a common validation pattern is unsafe, provide controls that expose the failure, and localize the bottleneck to a specific interface between graph learning and decoder-induced influence.

\section{Limitations}

This work does not solve causal attribution in observational MMM. CIG and AR-CIG evaluate a trained decoder's perturbation sensitivity; they are not interventional causal estimands without additional assumptions such as adequate temporal variation, measured confounding, stable response mechanisms, and plausible intervention support. Real marketing systems may contain hidden confounders, nonstationary platform shocks, correlated budget rules, delayed effects, and campaign-level constraints that are not fully represented by the synthetic generator.

The most important method limitation is exposed by our own experiments. The learned DICE posterior contains graph-recovery signal, but current graph interfaces do not reliably turn that signal into sparse decoder-consumable attribution support. Raw and hard interfaces become dense. Validation-MSE support selection fails. Stability selection remains weak. Even true-density top-$k$ gives only modest AR-CIG in the new sparse-target benchmark. A deployable attribution method needs label-free support selection or training objectives that directly improve calibrated sparse target-relevant graph support.

The baseline set is also intentionally focused. We compare to CausalMMM because it is the closest graph-MMM architecture and shares the graph-VAE interface. Broader temporal causal discovery baselines such as PCMCI, DYNOTEARS, and Neural Granger methods are relevant for graph recovery but do not by themselves test graph-mediated decoder attribution. Future work should include those baselines for graph recovery while preserving the decoder-bypass diagnostics introduced here.

Finally, the synthetic design provides known graphs but cannot exhaust real MMM behavior. The sparse-target benchmark removes the original target-parent degeneracy, but future studies should include richer mediated effects, budget constraints, paired generator counterfactuals, mROI ranking, budget-regret metrics, reversed-graph controls, degree-preserving random graphs, and out-of-distribution perturbation checks.

\section{Conclusion}

Our experiments show that attribution bypass is not merely a theoretical concern. In sparse-target MMM benchmarks, no-graph and full-graph decoders achieve nearly identical forecasting error to oracle-graph decoders, yet their AR-CIG alignment is near chance. Oracle graph support yields high AR-CIG, and frozen graph-swap shows that the same decoder can produce graph-aligned influence when supplied with the correct graph. Thus, the diagnostic and graph-safe decoder are functional. However, learned graph interfaces, including validation-MSE sparsification and stability selection, remain insufficient. This localizes the remaining challenge to deployable graph-support selection rather than forecasting or decoder capacity.

The paper's claim is therefore deliberately bounded. DICE-MMM is not presented as a final causal attribution engine. It is a diagnostic framework for exposing and localizing attribution bypass in graph-based neural MMM. Its main scientific value is to prevent low forecasting error from being mistaken for attribution faithfulness.

\section*{Reproducibility Statement}

All main results are reported as mean $\pm$ standard deviation over pre-specified seeds. We use the same generated datasets, splits, and evaluation protocol across methods. The graph convention is $A_{ij}=1$ when source $i$ influences receiver $j$; self-links are excluded from graph recovery and AR-CIG edge rankings. Graph-recovery sweeps follow the $R/d/T$ settings in Table~\ref{tab:graph_recovery}. The original response-diagnostic setting uses $R=10,d=10,T=120$ with 11 nodes including the target node. The sparse-target benchmark uses a non-degenerate target-parent structure. Stage-1 graph checkpoints are selected by validation graph diagnostics or the same late-window rule; Stage-2 decoder checkpoints are selected by validation target-horizon MSE. Graph-swap experiments freeze decoder weights and change only the graph input. CIG/AR-CIG perturbations use source replacement modes described in Section~3.4, with AR-CIG as the main attribution diagnostic. All experiments were conducted using two NVIDIA GeForce RTX 3090 Ti GPUs.

\subsubsection*{Acknowledgments}
This work was conducted while Yunbo Wang was an intern at Adsgency AI.

\bibliographystyle{plainnat}
\bibliography{main}





\end{document}